# A Comprehensive Guide to Enhancing Antibiotic Discovery Using Machine Learning Derived Bio-computation

Khartik Uppalapati[a], Eeshan Dandamudi [a], S. Nick Ice[a], Gaurav Chandra[a], Kirsten Bischof[a], Christian L. Lorson[a,b], and Kamal Singh[a,b,*]

[a] Bond Life Sciences Center, and [b] Department of Veterinary Pathobiology, College of Veterinary Medicine, University of Missouri, Columbia, Missouri, USA

¶Correspondence:

Kamal Singh, Ph.D.

440e Bond Life Sciences Center

University of Missouri,

Columbia MO 65211

(+1) 573-882-9024

ksingh@missouri.edu

**Abstract**

Traditional drug discovery is a long, expensive, and complex process. Advances in Artificial Intelligence (AI) and Machine Learning (ML) are beginning to change this narrative. Here, we provide a comprehensive overview of different AI and ML tools that can be used to streamline and accelerate the drug discovery process. By using data sets to train ML algorithms, it is possible to discover drugs or drug-like compounds relatively quickly, and efficiently. Additionally, we address limitations in AI-based drug discovery and development, including the scarcity of high-quality data to train AI models and ethical considerations. The growing impact of AI on the pharmaceutical industry is also highlighted. Finally, we discuss how AI and ML can expedite the discovery of new antibiotics to combat the problem of worldwide antimicrobial resistance (AMR).

1. **Introduction**

Traditional drug discovery processes are typically associated with challenges such as high costs, and extended timelines to approval. On average, bringing a new drug to the approval stage using traditional drug-discovery approaches takes nearly a decade of rigorous *in vitro*, *in vivo* assays, and clinical trials, costing nearly $3 billion per drug (DiMasi et al., 2016). This lengthy and costly processes is also riddled with failures as many potential drugs fail at different stages of the clinical trials, further complicating the process and cost of drug development.

Machine Learning (ML), a part of Artificial Intelligence (AI), has emerged as a promising tool in drug discovery as it can counter the challenges faced in traditional drug-discovery processes (Dara et al., 2022; Stokes et al., 2020). ML is used to create algorithms after training from large and vast datasets to make predictions. In drug discovery, ML scans through biological data to identify patterns that are likely missed by humans. ML models can be used to predict activity and potential side effects of hit (or lead) compounds and can suggest new scaffolds with promising properties (Vamathevan et al., 2019).

The use of computational tools in drug discovery is not new. However, ML has significantly enhanced the capacity of Computer-Aided Drug Discovery (CADD). In the past, researchers heavily relied on the Quantitative Structure-Activity Relationship (QSAR) models to improve upon the potency of a 'hit' or a 'lead' compound. QSAR, which uses statistical methods to predict compound activity based on molecular structure, are powerful but can be too simplistic to capture intricate relationships needed for accurate predictions (Lavecchia, 2015), and therefore, QSAR does not work for all potential therapeutics.

Deep Learning (DL), a subset of ML, that utilizes neural networks with multiple layers is proving to be a powerful tool in drug discovery. DL models can automatically identify relevant features from the raw data. Thus, DL models are extremely useful for analyzing molecular structures and predicting interaction of a new drug(s) with the target proteins (Zhavoronkov et al., 2019). Multiple factors such as improvements in computational power, availability of large datasets, and the creation of new algorithms have strengthened DL tools. For example, the Convolutional Neural Networks (CNNs) have become increasingly popular for analyzing molecular structures and predicting the affinity of a drug with its target (Gawehn et al., 2016). Similarly, the generative models can be used to design entirely new molecules with desired properties, transforming the landscape of drug discovery (Segler et al., 2018; B. Zhang et al., 2023).

This review is divided into the following sections. First, we explore the evolution of ML techniques (Section 2) followed by a focus on emerging models and techniques such as Graph Neural Networks (GNNs) and transformer models within the context of drug design and application of AI to antimicrobial resistance (AMR) (Section 3). While addressing the application of AI in drug discovery associated in AMR, we discuss global economical and clinical issues. Of note, AMR is a serious global public health problem, which threatens treatments of a variety of infections such as *Staphylococcus aureus*, *Klebsiella pneumoniae*, non-typhoidal *Salmonella* and *Mycobacterium tuberculosis* (Sakagianni et al., 2023). ML methods have been developed to analyze bacterial genomes, recognize epidemic patterns, forecast susceptibility to medication, and propose new antibacterial treatments (Sakagianni et al., 2023). ML methods can diagnose AMR rapidly (0.5-3 hours) (Lechowicz et al., 2013), thereby significantly aiding in the treatment regimen design and implementation. We then examine recent advancements in data-driven drug discovery highlighting drug discovery against cancer, and include the use of AI for high-throughput screening (Section 4). Then the ethical and regulatory considerations associated with AI in drug discovery are addressed (Section 5). In Section 6, we address current challenges in AI-based drug discovery. We then dive deeper into the challenges of AI integration as well as future trends in the field (Section 7). Finally ending with reflection on the potential of AI to affect the pharmaceutical industry (Section 8).

2. **Evolution of Machine Learning Techniques in Drug Discovery**

2.1. *Traditional Algorithms to Advanced Deep Learning.*

AI-based drug discovery has evolved significantly. Initially, traditional algorithms such as support vector machine (SVM) and Random Forest (RF) dominated the field as they were instrumental in the early stages of ML application, particularly in tasks like QSAR modeling. QSAR modeling helped predict the biological activity of chemical compounds based on their molecular structure. SVMs, on the other hand, excelled in classifying compounds by drawing optimal boundaries between different classes in high-dimensional spaces (Jaganathan et al., 2021; Lee et al., 2017). RF models provided robustness through its ensemble learning approach in which multiple decision trees were combined to improve prediction accuracy while also reducing the risk of overfitting (Cortés-Ciriano, 2015; Coudray et al., 2018).

1.a.1 *Limitations of Traditional Algorithms.*

While traditional ML algorithms were successful initially, the complexity and volume of biological data exposed the limitations of traditional ML algorithms. For example, when SVMs were applied to larger datasets common in drug discovery the models were overwhelmed and accuracy issues arose in the prediction process (Syed et al., 2020). RF models were more robust in conditions of overfitting and provided better generalization, but they also struggled to handle complex data relationships and therefore required extensive tuning in order to optimize performance (Elbadawi et al., 2021; Gupta et al., 2021).

### 1.1.2. *Evolution of Deep Learning.*

The introduction of DL marked a critical point in the use of ML in drug discovery as DL models such as CNNs and Recurrent Neural Networks (RNNs) outperformed traditional ML algorithms, primarily due to their ability to automatically extract salient features from the data sets without the requirement of manually extracting the features for use as in traditional AI algorithms (Jimenez et al., 2018; Zhavoronkov et al., 2019). CNNs, which were originally designed for image recognition tasks proved to be useful in analyzing molecular structures as well as predicting protein-ligand interactions. CNNs were able to offer higher accuracy due to their ability to capture spatial hierarchies in data (Gupta et al., 2021). RNNs, specifically Long Short-Term Memory (LSTM) networks, were initially created to handle sequence prediction tasks like predicting the effects of drug interactions over a period of time, making them extremely useful in comprehending temporal dynamics in biological systems (Choudhury et al., 2022).

With the evolution of ML models, the datasets used to train and accompany these models have also expanded. Datasets used in the most recent drug discovery ML models are shown below in **Table 1.**

### 1.1.3. *Generative models and drug design.*

Looking at past DL models, generative models have also emerged as powerful tools in drug design because these models, such as Generative Adversarial Networks (GANs) and Variational Autoencoders (VAEs), have the unique ability to generate new data instances like novel chemical structures that could likely serve as drug candidates (Chaudhari et al., 2020; Madeira et al., 2019). GANs, for example, consist of two neural networks, the generator and the discriminator, which leads to the creation of highly realistic synthetic data that can simulate the drug discovery process. This approach has been especially useful when applied for the expansion of chemical libraries with novel compounds that otherwise would not have been considered if traditional methods were used (Prasad and Kumar, 2021).

2. **Machine Learning Techniques in Drug Discovery and Application to AMR**

2.1. *Reinforcement Learning for Drug Development.*

Reinforcement Learning (RL) is a ML technique that learns from doing. Each action such as adding an atom or bond to a molecule earns a reward or penalty and helps the model to learn choices leading to the best outcomes. This is particularly useful in drug design where one tries to find the optimal structure for a molecule that would likely provide desired biological activity (Popova et al., 2018; Zhou et al., 2019). When it comes to targeting multiple factors (e.g. proteins) at once for drug design, RL is particularly powerful, which makes it critical for treating complex diseases like cancer. For instance, the MTMol-GPT model uses RL to generate molecules that are precisely tailored to effectively target and engage multiple molecules (Agyemang et al., 2021; Ho and Ermon, 2016). However, RL does come with its own challenges, like defining the right reward parameters (*e.g.,* potency, selectively, and safety) and managing the high computational costs. These reward functions need to be carefully fine-tuned so that the model focuses on the right balance of characteristics instead of becoming fixated on certain aspects. Another challenge is the sheer computational power needed for RL. Exploring the vast space of possible molecular structures and training the models that are useful requires significant processing power, which can be a major roadblock for many researchers (Popova et al., 2018; Zhou et al., 2019). Importantly, with the ongoing advances in algorithms and technology, RL is expected to play a larger role in the future of drug design processes (Popova et al., 2018; Zhou et al., 2019).

2.2. *ML in Cheminformatics.*

ML has grown to be an integral part of cheminformatics, offering advanced tools for the analysis and prediction of chemical properties, reactions, and biological activities (Chen et al., 2018). One of the most important applications of ML in cheminformatics is the prediction of molecular properties and activities. ML algorithms can be trained with very large data sets of chemical compounds to predict the physicochemical properties of new molecules such as efficacy, solubility, permeability, and toxicity (Paul et al., 2021). These properties can be successfully predicted by ML models to a high degree of accuracy, outperforming traditional QSAR models (Chen et al., 2018).

Another critical application of ML in cheminformatics is the design and optimization of new chemical compounds. Generative models, like VAs and GANs are increasingly being applied in the design of new molecules with specified properties (Sanchez-Lengeling and Aspuru-Guzik, 2018). Such models can generate chemical structures that are likely to possess specific biological activities, thereby fast-tracking the drug discovery processes. Moreover, reaction conditions in chemical synthesis can also be optimized by ML algorithms, hence making the manufacturing process more efficient and sustainable (Segler et al., 2018). These developments demonstrate the potential of ML to transform progress in the field of cheminformatics and its applications in the processes of drug discovery and chemical research.

2.3. *Transformer Models in Drug Design.*

Transformer models, whose initial purpose was for language processing, are now becoming useful in drug design. These models treat molecular structures as sentences by breaking them into sequences [similar to SMILES (Simplified Molecular Input Line Entry System) strings] and then using advanced language processing techniques to analyze them. This approach is particularly powerful when it comes to designing new drugs from scratch because transformers can generate entirely new molecular structures that meet specific design criteria (Vaswani et al., 2017(Popova et al., 2018).

A critical component of transformer models is their ability to deal with complex relationships in a molecule, even when the atoms are far apart from each other. This makes transformer models extremely useful at modeling large, intricate molecules that are common in drug discovery. However, transformer models require extensive computational power and large datasets to function effectively, which continues to be a challenge, but when combined with other models such as GANs, transformer models can create new drug-like molecules with specific properties (Olivecrona et al., 2017; Popova et al., 2018).

2.4. *Graph Neural Networks (GNNs).*

Due to its ability to model complex relationships within molecules, GNNs have quickly become a necessary tool in drug discovery and are the newest representatives of ML techniques for drug discovery. GNNs work by imagining each atom as a point (or node) and each bond as a line (or edge) that connects these points and using this structure to understand how a molecule's parts interact with each other. This, in turn, helps predict how a drug will behave in the body. Unlike older models that relied on fixed rules, GNNs learn directly from these molecular graphs which make them much better at capturing the unique characteristics between different molecules (Hamilton, 2020; Wang et. al, 2023).

GNNs' ability to combine information makes them extremely unique. For example, when a GNN analyzes a molecule, it starts by looking at each atom and its immediate connections. From that, it then updates this information layer by layer until it finally builds a complete picture of the molecule. This layered approach helps the model understand the small details along with the bigger picture which helps create more accurate predictions, regardless of whether trying to understand how a drug binds to a protein or how it might be metabolized (Kovacs et al., 2019; Veličković, 2018).

Another useful feature of GNNs is its ability to use attention mechanisms which allows the model to focus on the parts of the molecule that are most relevant to a given task like identifying a potential binding site for a drug. This is especially useful when dealing with complex or unusual molecules where some parts of the structure might be more important than others (Sanchez-Martin et. al. 2024; Yao et al., 2024)). Compared to an older model, GNNs offer a better dynamic and flexible way to model molecules, which is extremely important for identifying new drug candidates (Kipf and Welling, 2016; Liu et al., 2020).

## 2.5. *AI-Assisted Molecular Docking*

There are many open source and commercial docking programs that are used for the docking of small molecules into the target proteins. The widely used open-source programs are AutoDock (Goodsell and Olson, 1990), AutoDock Vina (Trott and Olson, 2010), and SMINA (a fork of AutoDock Vina) (Koes et al., 2013).

AI-assisted docking is a significant development in the drug discovery process as it improves upon traditional docking methods such as AutoDock and GLIDE. While traditional docking remains an invaluable methodology, it is computationally expensive and time-consuming process due to its reliance on physical simulations and scoring functions. AI-assisted docking can improve upon this process by training models on large datasets of known protein-ligand complexes that allow better predict binding outcomes in a short timeframe (Gainza et al., 2020).

Since AI-assisted docking depends on learning from the large datasets of known protein-ligand complexes, ML models such as GNN can predict the most likely binding poses and affinities. AI-assisted docking is especially useful when screening large libraries of compounds against a given target protein as it significantly speeds up the early stages of drug discovery (Gligorijevic et al., 2021; Jumper et al., 2021). AI models add additional layers of information such as molecular dynamics with docking, which can provide more detailed information on the flexibility of molecules over a period of time. This, in turn, improves docking predictions and makes them more robust. An example of this involves the highly accurate prediction of binding affinities as even difficult targets like protein-protein interactions or highly flexible binding sites to a high degree of accuracy because of AI techniques. While these AI-driven approaches save time, they also pave the way for new processes to discover drugs that might have otherwise been overlooked (Gainza et al., 2020) Wallach et al., 2023).

Newer and faster AI-based molecular docking programs are continuously emerging. For example, GNINA (McNutt et al., 2021) utilizes an ensemble of CNNs as a scoring function outperforms AutoDock Vina scoring on redocking and cross-docking tasks when the binding pocket is defined. Other ML based docking programs include DiffSBDD, developed at Massachusetts Institute of Technology and now adopted by NVIDIA, uses a three 3D equivariant GNN that has three layers: embedding, interaction layer with 6 graph convolution layers, and output layer (Schneuing et. al, 2022). Additional examples of AI-based molecular docking programs are EquiBind (Li et al., 2023), TankBind (Lu et. al, 2024), and PoseBusters (Buttenschoen et al., 2024). An advantage of these molecular docking programs is that they are available through open-source licenses as opposed to commercial docking programs such as Glide from Schrödinger Inc. (Bhachoo and Beuming, 2017; Friesner et al., 2004; Morris et al., 2009) and GOLD (Verdonk et al., 2003). A list of opensource molecular docking programs can be found at https://biochemia.uwm.edu.pl/en/docking-2/. The efficiency of opensource docking programs can be enhanced by combining the opensource binding site prediction programs such as P2Rank (Krivak and Hoksza, 2018), AutoSite (Ravindranath and Sanner, 2016) and AutoLigand (Harris et al., 2008) among others.

### 2.6. QSAR Methods for Antimicrobial Activity.

With the rapid evolution of antimicrobial resistant pathogens that outpace traditional drug discovery methods QSAR could offer a potential solution.. For this purpose, ML has become important, especially through the use of QSAR modeling. With this technique, researchers can predict the efficacy of potential antimicrobial compounds by analyzing the relationship between their molecular structure and biological activity.

1.a.1 *QSAR Modeling in Antimicrobial Discovery.*

QSAR is an important technique to discover new antimicrobials. QSAR functions by examining the structural properties of chemical compounds such as molecular weight, hydrophobicity, and electronic characteristics. QSAR will then correlate these properties with the biological activity of the compounds. By using QSAR models, researchers have been able to predict which new compounds possess better efficacy and show effective antimicrobial activity, thereby potentially reducing the time and cost associated with traditional drug discovery (Bento et al., 2014).

A notable use of QSAR modeling to fight against multidrug resistant bacteria comes from the work of Bento and colleagues in 2015 as they focused on using QSAR models to screen and predict antimicrobial activity for a wide array of chemical compounds. By delving deep into the structure-activity relationships (SAR) of existing antibiotics, they were able to identify patterns and molecular characteristics that shared a correlation to strong antimicrobial properties. They did so by attaching QSAR models to a large dataset of antibiotic compounds in order to thoroughly analyze how specific molecular features influenced the compounds' ability to combat resistant bacteria. Their models did not simply identify potential candidates from pre-existing molecules, rather, they were able to successfully predict several novel compounds that subsequently exhibited promising antimicrobial activities. These compounds were then experimented on further for validation which highlights how computational models can complement traditional methods to accelerate the drug discovery process while also reducing costs and resource use (Bento et al., 2014).

One issue with QSAR modeling in antimicrobial activity is its reliance on the availability of high quality, experimental data. QSAR models require extensive datasets of chemical structures as well as their corresponding biological activities to accurately predict the antimicrobial potential of new compounds. If the data used is incomplete, biased, or not representative of the full chemical space, the model could become overfitted and lack generalizability which will make it less reliable when introduced to new data (Cherkasov et al., 2014).

2.6.2. *Integration of QSAR with Advanced ML Algorithms.*

The integration of advanced ML models with QSAR modeling has shown to improve the prediction of antimicrobial activity. The ML-infused QSAR models can capture the intricate as well as non-linear relationships between chemical structures and their biological activities against pathogens. For instance, DL algorithms such as neural networks are especially good at handling complexities in antimicrobial drug discovery since they learn from extensive datasets to identify subtle patterns and correlations that traditional QSAR methods may miss. This is especially true in the situations where the chemical structures are unconventional or where the biological data is heterogenous or incomplete (Bento et al., 2014; Cherkasov et al., 2014)).

When applying ML-enhanced QSAR to antimicrobial activity, models can more accurately predict the efficacy of novel compounds against resistant bacteria, such as methicillin-resistant *Staphylococcus aureus* (MRSA) or multi-drug-resistant *Pseudomonas aeruginosa* (Koh et al., 2023). Additionally, these models can quickly identify off-target binding of novel compounds causing toxicities. ML tools contribute heavily to improving the predictive power of QSAR models which therefore allows it to quicker and more efficiently create new antimicrobial agents (F. Zhang et al., 2023).

2.7. *The Use of ML in Antimicrobial Resistance Prediction.*

The early identification and prediction of AMR can help avoid its economical and clinical drawbacks. Thus, genome-based ML can be applied to AMR prediction, using a data- driven approach rather than a rule-based prediction (Benefo et al., 2024). For example, genomic information on isolates associated to the specific disease can be combined with data on these isolates' AMR phenotypes to different antibiotics. Commonly, these include amoxicillin-clavulanic acid, ampicillin, ceftiofur, ceftriaxone, sulfisoxazole, streptomycin, tetracycline, and cefoxitin (Benefo et al., 2024).

The best algorithms for genome-based prediction of AMR are RF, Cost-Sensitive Learning, SVM, Radial Bias Function Kernel, Neural Network, and Logit Boost. These algorithms are trained on data specific to the disease and can be evaluated for the model fit based on the Area Under the Receiver Operating Characteristic curve and confusion matrix statistics. For each antimicrobial, the model can predict the AMR phenotypes of the set of isolates, which can be compared to ResFinder AMR phenotype predictions to test accuracy (Benefo et al., 2024; Florensa et al., 2022).

Finally, AI assists Genome Wide Association Studies (GWAS), which can test the association between genetic variants and corresponding AMR phenotypes to find new resistance determinants. The results of GWAS are broad ranging, as they can gain insight into a phenotype's underlying biology, predict a phenotype's heritability, calculate genetic correlations, and more (Uffelmann et al., 2021). Crucially, GWAS can inform drug development programs to develop more effective antibiotics.

Specifically, ML models can use Genome Scale Models (GEMs) for high-performance classifying of AMR phenotypes of disease strains. GEMs can directly use GEMs as an input-output ML model to extract insights from GWAS datasets (Kavvas et al., 2020). Although traditionally using GWAS can have reduced prediction accuracy, the integration of ML can increase these rates.

3. **ML Applications in Phases of Drug Discovery – Cancer as an Example**

   3.1. *Joint Forces: Collaborations Between Pharmaceutical and Tech Companies.*

The rise of AI in drug discovery has spurred numerous collaborations between pharmaceutical companies and AI technology firms. Partnerships bring together the forte of both sectors: pharmaceutical expertise in drug development and technological might in AI and computational modeling.

   1.1.1. *Notable Pharmaceutical Collaborations.*

One example of this collaboration is when AstraZeneca joined forces with a company called BenevolentAI. This collaboration is focused on using AI to accelerate the drug discovery process by finding new therapeutic targets. Benevolent AI's platform brings ML together with enormous amounts of biomedical data to identify new drug candidates for complex diseases. This deal epitomizes the potential for AI to enhance traditional drug discovery processes and really accelerate the development of new therapies. AstraZeneca and BenevolentAI's collaboration is particularly notable for its focus on using AI to address areas of high unmet medical need, such as neurodegenerative diseases and autoimmune disorders (Brown et. al, 2018).

Another notable collaboration is between Pfizer and IBM Watson. The latter's AI capabilities were expected to be used in the partnership in analyzing huge datasets to identify prospective drug candidates. The AI platform of IBM Watson supported decision-making in clinical trials and helped increase the efficiency and accuracy of the design and execution of trials. This clearly showed how AI, while traditionally used in early stages of drug discovery, can also be used in optimizing later stages of drug development. Integrating AI into the clinical trial process has enabled Pfizer to structure the development of new drugs much better and bring drugs to the market at speeds that were previously impossible (Eyre et al., 2021).

The collaboration between Novartis and Microsoft is another success story in AI-driven drug discovery. Novartis harnessed the power of Microsoft's AI platform to accelerate the design of new drugs by analyzing heaps of biological data. This partnership was oriented toward the application of AI in developing new treatments against conditions with high unmet medical needs, like cancer and autoimmune diseases. If this collaboration worked, then this would go a long way to confirm the potential of AI to be an innovator in drug discovery, particularly for areas where more traditional methods of discovery have been found lacking. The Novartis-Microsoft deal exemplifies another critical aspect: state-of-the-art technology is required to be competitive in the pharma market today (Eichler et al., 2022). The Exscientia-Sanofi collaboration is another example. In the case of Sanofi, Exscientia was an AI-driven drug discovery company that collaborated for the development of novel small-molecule drugs. The AI platform of Exscientia was used to design and optimize drug candidates. This saved time and cost associated with drug discovery through conventional methods. It realized the potential for AI to fast-track the process of drug discovery by identifying a number of potential drug candidates. The success of this partnership marked another trend whereby pharmaceutical companies seek partnerships with AI firms to fast-track their drug discovery capabilities (Sanofi, 2022). Such collaborations between pharmaceutical companies and tech firms, therefore, become emblematic of the potential for AI to truly transform drug discovery. The way these partnerships can take the best of both worlds to drive innovation toward the successful development of new therapies and bring them to market faster works effectively at the interface between both parties.

### 1.1.2. *Successful Case Studies Resulting from Company Collaboration.*

Besides the above-mentioned collaborations, other major partnerships have significantly contributed towards AI-driven drug discovery. The GSK/Exscientia collaboration has been very productive in hastening drug discovery with AI. This partnership has resulted in the identification of several potential drug candidates, amongst which includes a molecule under clinical investigation against Chronic Obstructive Pulmonary Disease (COPD). In this regard, the very fact of the success of this partnership proves that AI holds huge potential not only in accelerating drug discovery but also enhancing the precision and effectiveness of the drug development process (Exscientia, 2023).

Another exemplary collaboration is between Merck and Atomwise. Atomwise's AI-driven platform uses DL for predicting the binding affinity between small molecules and target proteins—one of the most crucial steps involved in drug discovery. Also, through the collaboration with Merck, several novel drug candidates have been identified. It evidences AI's potentiation in the drug discovery process. This highlights how AI can blend into the traditional drug discovery workflow in its ability to increase the efficiency and effectiveness of the process (Peters et al., 2020).

Case studies such as these epitomize how AI-driven collaborations can really disrupt drug discovery. The potential for two-way collaboration by pharmaceuticals and AI technology firms to drive innovation and help bring new therapies to market more quickly and efficiently cannot be denied. However, their success is based on both parties' ability to cooperate and integrate their respective expertise and technologies.

### 1.2. Generating Synthetic Data with Artificial Intelligence.

Synthetic data is biological data generated by AI models based on real world data samples. Oftentimes with real data, scientists face issues such as poor-quality data, under-fitting of models, and barriers to data access, highlighting the importance of AI- generated synthetic data (Peters et al., 2020). Thus, synthetic data allows researchers to simulate different biological scenarios for therapeutic purposes without facing these challenges (Pun et al., 2023).

Models to generate synthetic data primarily feature GANs, which projects features onto an image space that the discriminator subsequently operates on, leading to Visual Domain Adaptation. In one case study, Microsoft researchers used ML systems based on a 3D face model and an asset library to generate synthetic data and found it had comparable accuracy to real data and detected faces in unconstrained settings (Lu et al., 2024). Aside from GANs, systems focusing on de novo drug molecular design have also used VAEs, energy-based models, diffusion models, reinforcement learning, and more (Lu et al., 2024).

Importantly, because it can augment or replace real data, synthetic data is also re-used to train ML models that again create synthetic cohorts. Specifically, synthetic data can be used to train AI models to identify therapeutic targets that may have been previously overlooked, allowing pharmaceutical researchers to cover possible holes (Pun et al., 2023).

An added benefit of AI-generated synthetic data is that it leads to better sample representation. It is common for certain patient populations to be underrepresented during data collection, leading to experimental bias. Fortunately, synthetic data can lead to more comprehensive and unbiased analyses by filling in data for these underrepresented groups (Chawla, 2002).

Unfortunately, there are limitations to using synthetic data that have yet to be addressed. Firstly, AI models cannot generate synthetic data with complexities that the model is not aware of, showing problems with cyclical training (Achuthan et al., 2022). Additionally, relying too much on synthetic data for underrepresented populations raises ethical concerns by completely forgoing relevant, real-world data (Howe and Elenberg, 2020). Therefore, although synthetic data has endless possibilities for scientific use, it should continue to be perfected with techniques like robust validation and quality control measures.

1.3. **Drug Target Identification and Validation.**

After the AI models have been trained using synthetic data and other large datasets of existing drug molecules, they can start identifying what processes or molecules (RNA, proteins, cellular pathways, etc.) should be targeted by a drug to prevent a certain disease (Qui et. al, 2024). This involves black box screening, which includes all types of phenotypic screening to determine which biomolecule leads to phenotypic changes that relate to disease. In technical terms, this entire process is known as target identification and is a crucial step of drug discovery. Challenges to traditional target identification are vast and include insufficient coverage of the patient population, lack of alternative responses to drug resistance, and the limited presence of "druggable" targets (You et al., 2022). In fact, out of all the potential "druggable" targets in the human genome, the majority have yet to be targeted, highlighting the importance of novel AI tools for optimization and efficiency (Verdine and Walensky, 2007).

### 1.3.1. *The Integration of Multi-Omics Data.*

The integration of multi-omics data has revolutionized drug discovery as it has provided a comprehensive view of biological systems. Multi-omics refers to the combination of various "omics" fields, such as genomics, proteomics, metabolomics, and transcriptomics, to analyze the molecular basis of diseases and drug actions more holistically (Cortés-Ciriano, 2015; Tian et al., 2013). By integrating data from different omics layers, researchers can identify correlations as well as causal relationships that might be missed when examining each data type separately. For example, the combination of genomic data (which provides insights into genetic predispositions) with proteomic data (which reveals protein expression levels) can help identify novel drug targets and understand disease mechanisms at a deeper level (Choudhury et al., 2022).

Many case studies show the successful integration of multi- omics into drug discovery. One instance is the successful integration of multi-omics in cancer research, in which multi-omics data was used to identify biomarkers for early detection in an attempt to develop personalized therapies. In one study, integrating genomic, transcriptomic, and proteomic data allowed researchers to predict patient responses to chemotherapy more accurately, allowing for more tailored treatment plans to be developed (Arora et al., 2021; Dara et al., 2022). Another example can be seen in the battle against drug resistance. By combining metabolomic and proteomic data, researchers have uncovered new ways that cancer cells react to chemotherapy to survive. Due to multi-omics however, strategies are now being developed to counteract this resistance and hopefully create more effective cancer therapies (Gupta et al., 2021).

Based on the importance of multi-omics data to drug discovery, AI should be used as a complementary and enhancement tool. For example, AI has already shown promise in integrating multi-omics data into networks that simulate pathways causing certain diseases (Bai et al., 2023). This integration also enhances the power of predictive drug discovery algorithms and the designs of patient-specific therapeutic regimens, speeding up drug discovery altogether (Khalifa and Albadawy, 2024). These predictive models are especially important as they identify a patient's risk to a disease before the condition develops, allowing time for early care (Pammi et al., 2023).

Another example is the Explainable Multi-Omics Graph Integration algorithm, which is a GNN algorithm that analyzed multi-omics pan-cancer data and identified 156 novel oncogenes that had the potential to become cancer targets (Guan and Wang, 2024). For overall central nervous system diseases, AI/ML tools were also powerful in identifying therapeutic targets after analyzing multi-omics data and ML (RF, SVM, and Decision Tree) classifiers predicted disease- associated genes using multi-omics data (Vatansever et al., 2021). Based on these previous studies, AI should further integrate with multi-omics data for different phases of drug discovery.

While integrating multi-omics data with drug discovery comes with many benefits, there are still many factors that need to be addressed for it to be fully utilized. One of the biggest issues is the size and complexity of the data that is being handled by the AI models (Elbadawi et al., 2021). It has also proven tricky to standardize data across different -omics platforms since each platform uses different techniques to gather their data. Due to this, a lot of variability happens between the data which makes it more difficult to handle and manage effectively both traditionally and with AI. To fix these shortcomings in the future, it is important that features be added to interpret the data and better ML models be created in order to take full advantage of multi-omics (Prasad and Kumar, 2021).

High-throughput technologies have generated an unprecedented amount and complexity of multi-omics data in the current scenario. AI/ML is a promising avenue for tapping the full potential of multi-omics data for drug discovery, as it can enhance the overall process. This will help speed up analysis by researchers or scientists and can be instrumental in deciphering difficult patterns and building predictive models that were not possible before. This acceleration in the processing of data results in very fast generations of testable hypotheses for enhanced drug discovery pipelines. Moreover, AI-driven pattern recognition can discover relationships between genomics, transcriptomics, proteomics, and metabolomics. Furthermore, AI is great at modeling the progression of diseases and, hence, predicting individual patient responses to various treatments – thereby giving an optimal selection of drug candidates.

### 1.3.2. *Identifying Disease-Associated Protein Inhibitors.*

A common way for drugs to treat disease is by targeting disease-associated proteins through inhibition. AI has been proven to identify potential inhibitors that act as bases for drug development. For example, one of the most striking successes of AI-driven drug discovery is the development of DDR1 kinase inhibitors. Researchers applied DL algorithms to identify potent DDR1 inhibitors and completed the discovery of a lead candidate in just 21 days. It involved generative AI, synthesis, and in vitro and *in vivo* testing of the compounds (Zhavoronkov et al., 2019). This example denotes that AI can significantly reduce timelines for drug discovery, which often reach years. Not only was the success of the study proof of the swiftness and efficiency of AI-driven approaches, but it also served as an example that the amalgamation of traditional empirical validation and AI-driven approaches can lead to a more streamlined process for discovering drug-like compounds.

One other large AI-driven effort is that concerning the MALT1 inhibitor SGR-1505, developed by Schrödinger. The campaign made a marriage of physics-based and ML approaches to computationally screen 8.2 billion compounds. Within this process, a clinical candidate was selected in just 10 months, with 78 molecules synthesized. This exemplifies the pace at which this process has been carried out and, by implication, the potential for AI to really transform the process of drug discovery. By drastically reducing the number of molecules that need to be prepared in the lab and subsequently tested, AI facilitates a much more focused and efficient approach toward drug development.

Furthermore, a drug lead discovery AI platform by Exscientia has successfully identified a novel CDK7 inhibitor that progressed into clinical studies. This was possible because AI can optimize the lead compounds through iterative cycles of prediction and validation quickly. Such an approach, according to Exscientia, could reduce the time taken for drug development to less than two years from target identification to clinical trials (Mullard, 2024).

### 1.3.3. *Synthesis Planning and Rewards-based Route Optimization Workflow (SPARROW).*

One recent development in data-driven target identification is Synthesis Planning and Rewards- based Route Optimization Workflow (SPARROW). Published just this year, SPARROW is a quantitative decision-making framework developed by MIT researchers Jenna Fromer and Connor Coley that automatically identifies optimal molecular candidates to test in new compounds. Alongside pinpointing molecules, the SPARROW algorithm provides a list of materials and an experimental procedure to synthesize the molecules (Fromer and Coley, 2024).

What sets SPARROW apart from other systems is the mathematical basis of its optimization, mirroring what is known as "expert intuition" when selecting synthetic routes (Fromer and Coley, 2024). The data that SPARROW sources from include virtual libraries, de novo design, and expert ideation produce a candidate set of molecules. Then, property prediction and retrosynthetic planning identify which molecule in the set performs best in a cost-benefit analysis. Finally, the algorithm specifies which candidates, and synthetic routes meet the cost-benefit standard, and this optimal subset of candidates is ultimately put forward for testing (Fromer and Coley, 2024).

One of the main graphs that is part of the SPARROW algorithm is the retrosynthetic graph, which is a bipartite graph relating reactants and products, which are the candidate molecules. This graph acts as a map to examine potential synthetic pathways for molecules and/or compounds (Gupta, 2024). Additionally, the SPARROW framework uses a linear optimization problem as part of linear optimization theory. This problem is solved with the CBC (Cutting plane branch and check) solver and the Python package PuLP (Gupta, 2024).

In terms of the specific equations SPARROW operates on, it uses both a nonlinear objective function and a simplified objective function. The nonlinear objective function maximizes the expected reward per unit cost of selected candidates whereas the simplified linear objective function shows the scalarized optimization of reward and cost. The specific mathematical equations SPARROW operates on are shown above in Tables 1 and 2. There are three constraints to SPARROW's problem formulation. First, if a reaction is selected, all reactants must also be selected. Second, if a compound node is selected, one or more of its parent reactions must be selected. Third and finally, for each cycle in the graph, every reaction node cannot be selected at the same time (Fromer and Coley, 2024). The expressions for these constraints are shown below.

**Nonlinear Objective function for SPARROW**

$$\text{argmax}_{c,r} \sum_{j \in \mathcal{T}} c_j U_j \prod_{i \in \mathcal{R}_j} L_i \text{cost}(\mathcal{R}_j | \forall j \in \mathcal{T} : c_j = 1)$$

**Simplified Linear Objective Function for SPARROW**

$$\text{argmin}_{c,r} \left( -\lambda_1 \sum_{j \in \mathcal{T}} c_j U_j + \lambda_2 \sum_{i \in \mathcal{S}} D_i r_i + \lambda_3 \sum_{i \in \mathcal{R}} \min\{L_i - 1, 20\} r_i \right)$$

**Constraint 1**

$c_j \geq r_i \quad \forall j \in \mathcal{P}_i, \, i \in \mathcal{R}$

**Constraint 2**

$\sum_{i \in \mathcal{P}_j} r_i \geq c_j \quad \forall j \in \mathcal{C}$

**Constraint 3**

$\sum_{i \in \mathcal{Y}} r_i \leq \text{length}(\mathcal{Y}) - 1 \quad \forall \mathcal{Y}$

Where, $j$ is an index referring to a reaction node; I is an index referring to a compound node; $R$ is a set of reaction node indices; $C$ is Set of compound node indices; $c_j$ Decision variable defining whether compound node $j$ is selected; $r_i$ is Decision variable defining whether reaction node $i$ is selected; $S$ is a set of dummy reaction node indices; $P_i$ or $_j$ is Set of parent nodes for the node corresponding to index $i$ or $j$, and $Y$ is a cycle in a retrosynthetic graph

Crucially, the use of AI to streamline drug discovery has proven costly with the price of materials and the required precise steps for synthesis that pose high risks. Even using substitution based on synthetic accessibility scores (SAscore, SYBA, SCScore, and RAscore) or general synthetic complexity has not been effective in significantly lowering the sum of costs (Gupta, 2024). As a result, SPARROW serves as a useful method for scientists to make cost- aware decisions and eventually lower already-high drug prices by using a multi-objective optimization criterion in the development process (Awadallah et al., 2023).

In three case studies based on real-world issues faced by chemists, SPARROW was successful in identifying potential molecular candidates, summarizing the costs of synthesis, and providing lists of procedures and materials (Fromer and Coley, 2024). SPARROW is still in development and supported by a variety of experts including Patrick Riley and John Chodera, Defense Advanced Research Projects Agency, the Office of Naval Research, and the National Science Foundation (Fromer and Coley, 2024). Clearly, there is broad interest in advancing AI applications for a variety of disciplines and SPARROW is one of the tools to achieve this goal.

1.3.4. *Cancer-Based Case Studies.*

In the case of cancer, AI allows scientists to identify novel anticancer targets and the novel drugs that will target them. Specifically, the process uses multi-omics data – epigenetics, genomics, proteomics, and metabolomics – and runs it through network based and ML-based models for application to target identification, drug screening, and more (You et al., 2022). One case study for target identification used weighted gene co-expression network analysis to analyze cancer data and found that certain hub genes and hub miRNAs were predictors of cancer in patients (Zhou et al., 2018). Generally, over thirty studies in the past seven years have focused on cancer and successfully used ML models to find prognostic and predictive cancer biomarkers (Al-Tashi et al., 2023).

The specific techniques used by these models are classification, clustering, and proteomics-based signaling pathway analysis. Classification helps sort potential biomarkers of defined classes based on microarray data and can be used to reveal subtypes of disease and their molecular characteristics. Once classification is done, clustering can be applied to proteomics data to identify clusters of samples that can be compared to previously reported disease subtypes. Finally, proteomics-based signaling pathway analysis distinguishes disease subtypes and finds specific pathways that may cause disease progression.

These techniques were applied to pancreatic cancer, and it was determined that perturbations in the SMAD4 or mTOR signaling pathways could be drivers of oncogenesis (Sinkala et al., 2020).

1.3.5. *Validation of Identified Targets.*

Once these targets are identified, they need to be validated, meaning the drugs must prove therapeutic benefits within an acceptable window. Target validation can involve determining structure–activity relationships, manipulations of target genes (knockdown or overexpression), monitoring signaling pathways downstream of the presumed target, and more (Visan and Negut, 2024). AI has already been used to validate targets by prioritizing the ones with the highest therapeutic potential and reducing false results, ensuring that researchers can be confident in their identified target (Pun et al., 2023).

However, some of these network-based biological analysis tools are not able to handle the onslaught of information from multi-omics data and therefore can lead to false positives (Bodein et al., 2022). Fortunately, continued innovation is likely to fix this issue and improve the precision of these ML-based systems (You et al., 2022).

1.4. ***Predictive Screening of Psychochemical Properties.***

For drugs to be successful, they need to meet desired psychochemical properties – bioactivity, toxicity, and aqueous solubility. Manually, it is time-consuming and costly to predict the effects of developed drugs using previously mentioned high-content screening. Fortunately, models based on AI and ML are effective in predicting these three qualities (Paul et al., 2021).

1.4.1. *Bioactivity.*

Predicting the bioactivity of drug molecules measures drug efficacy by testing Drug Target Binding Affinity (DTBA) for a target protein. The better a drug can interact with a target protein, the better it can help elicit a therapeutic response. By considering the chemical features or similarities of the drug and its target, AI-driven methods can predict and measure the DTBA of a drug (Paul et al., 2021). AI-based tools such as XenoSite, FAME, and SMARTCyp determine the metabolism sites of the drug while CypRules, MetaSite, MetaPred, SMARTCyp, and WhichCyp can identify specific isoforms that mediate a certain drug metabolism (Paul et al., 2021). In one study that accounted for 318 individual projects, a CNN titled AtomNet selected novel molecules across every therapeutic and protein class without known binders, X-ray structures, or manual cherry-picking (The Atomwise AIMS Program, 2024) (Atomwise, 2024). In 91% of the experiments, AtomNet was able to identify single-dose hits that were later reconfirmed in dose–response experiments. Specifically, there was a 74% hit rate for protein-protein interaction sites and a 79% hit rate for allosteric sites (Atomwise, 2024).

1.4.2. *Toxicity.*

Predicting the toxicity of a drug directly relates to human health and ensures the safety of products. Current *in vitro* and *in vivo* toxicity tests pose the same challenges as previously mentioned whereas computational tools help evaluate drug toxicity accurately and economically (Zhang et al., 2018). Generally, AI-based toxicity predictions use fast supervised models, which include RF, deep and graphical neural networks (Tonoyan, 2024). An example of these supervised learning algorithms is the ADME-Tox (absorption, distribution, metabolism, excretion and toxicity) Predictor, which uses the existence of compounds with known pharmacokinetic properties to predict the toxicity of early-stage drugs (Maltarollo et al., 2015). Another ML tool for prediction of toxicity is DeepTox (Mayr, 2016), which outperformed traditional methods and used 2500 predefined toxicophore features alongside molecular weight and Van der Waals volume (Paul et al., 2021). In 2023, when researchers were aiming to develop novel antibiotics for methicillin-resistant *Staphylococcus aureus* (MRSA), they trained three D models to predict whether a list of 12 million identified compounds were toxic to different human cells (Wong et al., 2024). More recently, the ChemTunes' ToxGPS platform combines ML with quantitative structure- toxicity relationships to assess a given compound's toxicity. It relies on numerous databases, including physicochemical parameters, xenobiotic metabolism, toxicokinetics, and the ToxCast/Tox21 database (Tonoyan and Siraki, 2024). Reducing the number of drugs with undesirable toxicity profiles is key to decrease clinical trial failures and harmful side effects of medications.

1.4.3. *Aqueous Solubility.*

Aqueous solubility is an important parameter to ensure drugs are absorbed properly and meet the desired concentration in systemic circulation. Without high solubility, drugs may not elicit the intended pharmacological response in the body (Savjani et al., 2012). Because of flaws in traditional predictive methods, more than 40% of drugs developed by the pharmaceutical industry are insoluble and widely ineffective (Kalepu and Nekkanti, 2015). On the other hand, ML approaches accurately predict compound solubility and shorten process times (Xue et al., 2024). From decision tree, RF, and gradient boosting algorithms, the latter was found to be the most effective in predicting drug solubility. Within gradient boosting algorithms such as LightGBM, CatBoost, and XGBoost, a study determined that CatBoost showed the greatest potential, with higher model complexity based on $R^2$ score and a smaller file size (Xue et al., 2024). In summary, AI-based methods reduce the reliance on traditional, more resource-intensive processes to predict the bioactivity, toxicity, and aqueous solubility of a new drug.

## 1.5. *Clinical Trials and Outcome Prediction.*

The final step before drugs are released as products is clinical trial testing, allowing researchers to determine whether or not drugs are safe and effective for people to use. Currently, pharmaceutical companies are forced to spend billions to fund clinical trials because of inefficiency and safety concerns (Hardman et al., 2023; Zhou et al., 2023).

However, much like SPARROW which was mentioned earlier, most AI models being used for drug discovery take information and extract it into procedures for clinical trials, eliminating the need for traditional methods (Singh et al., 2023). Specifically, AI is able to look at thousands of previous studies to not only recommend the most efficient design, but also which patients may be the best fit for the study (Liu et al., 2024). The procedures can generally be broken down into a few categories such as what dosage should be given, how many patients should be recruited, and what data should be collected from the patients to name a few (Fu et al., 2022).

At the University of Illinois Urbana Champaign (UIUC), Jimeng Sun lab has developed a hierarchical interaction network (HINT) (Fu et al., 2022) that can predict the success of a trial based on the drug, disease, and patient recruitment criteria. Based on the result, pharmaceutical companies can either push forward with the trial or make revisions to the trial design (Fu et al., 2022). Researchers at Stanford Medicine and McMaster University also developed SyntheMol, which generates recipes for chemists to synthesize antibiotics in their lab. In just one run, the model generated synthesization procedures for six novel drugs to kill resistant strains of *Acinetobacter baumannii* (Swanson et al., 2024).

Stanford University researchers additionally developed the CliniDigest tool, which uses GPT-3.5 to summarize batch clinical trials. Specifically, CliniDigest is able to reduce 10,500-word descriptions into concise 200-word summaries that include in-text reference citations (White et al., 2023). This system is important to create simpler descriptions of studies when pharmaceutical researchers are looking for references for their own clinical trials. From start to finish, AI plays an important role in improving efficacy for these clinical trials and will continue to save companies time and money in the future.

**Ethical and Regulatory Considerations in AI-Driven Drug Discovery**

The adoption of generative AI in the pharmaceutical industry is relatively new and marked by several regulations. Despite its immense potential, it is important to note that there are several ethical and regulatory limitations on AI-driven drug discovery and the broader new technical capabilities being developed.

Because AI is only able to operate based on the input data it is trained with, biased and untransparent datasets can risk skewed outcomes. Additionally, there is a high likelihood that AI outputs will only replicate these biases (Currie et al., 2024). For instance, training data skewed towards certain demographics can result in incorrect treatment and diagnosis for these groups when the opposite should be occurring (Delgado et al., 2022). Some companies have raised questions as to when and how patients are informed about the role of technology in their treatment, trials, or diagnoses (Rosa et al., 2021). The only sustainable solution to this issue is to have individuals and companies themselves create strategies that integrate fairness into their therapeutic solutions.

Biases can also impact clinical trials and drug development, as the protocol designs discussed earlier may show incorrect and unrepresentative instructions for patient recruitment and data analysis, compounding existing socioeconomic inequities. When it comes to trial participants, diversity is key in understanding how drugs impact different racial, gender, and demographic groups (Dwivedi et al., 2023). When the design of a clinical trial itself is flawed, it leads to devalued research publications. Suggestions to fix this issue include data safety monitoring boards and mandates on clinical trial result publications, allowing companies to prioritize patient wellbeing (Van Norman, 2021). Furthermore, the synthetic data mentioned in Section 5 can be used for data augmentation to increase diversity in datasets (Blanco-Gonzalez et al., 2023).

Additionally, the results that AI predicts for clinical trials can be hard to reproduce and require large amounts of training data, creating concerns for security and patient privacy (Fleurence, 2024). The European Union has already taken a few steps to address these regulatory concerns, including requiring AI and ML technology to comply with existing standards such as GxP standards on data and algorithm governance and nonclinical development principles (European Medicines Agency, 2023). Moreover, cost-benefit analyses of these models must feature a "risk-based approach" (European Medicines Agency, 2023).

Aside from concerns for clinical trials, there is also criticism about the use of AI in the pharmaceutical industry causing job loss due to automation (Blanco-Gonzalez et al., 2023). However, some also say that AI could supercharge higher-order hiring in other sectors of the economy to offset AI-related job losses. Historically, there was a trend of increased hiring following rises in automation across different industries (Webb, 2019). November election and Federal Reserve decisions for interest rates, will decide the pharmaceutical industry related job creations (Webb, 2019).

In the United States, the Food and Drug Administration (FDA) recognizes the increased use of AI models in drug discovery and is working with the Center for Biologics Evaluation and Research and the Center for Devices and Radiological Health to create an agile regulatory ecosystem that balances innovation with public health. The number of submissions for drug applications has also increased significantly, increasing FDA engagement in this area and supporting partnerships with other organizations globally (Takebe et al., 2018).

Although expecting completely perfect systems before implementation is unrealistic, steps should still be taken to ensure AI models can address ethical and regulatory challenges. Utilizing the suggestions mentioned in this section alongside new proposed developments will allow companies to meet these goals.

**Challenges in AI-based Drug Discovery**

AI-driven drug discovery has made impressive progress but is riddled with a long list of problems that have limited its potential, from data quality and model generalization to how ML can be seamlessly integrated into existing drug discovery pipelines (Chan et al., 2019). Next-generation trends in AI—from Quantum Computing to Federated Learning, and finally to AI for personalized medicine—come with new opportunities and challenges (Gorgulla et al., 2020). Aside from technological advancements, the field has been showing increased popularity amongst researchers and scientists.

As shown in Figure 1, there has been a near-exponential increase in the number of articles discussing the use of AI in antibiotic drug discovery over the past five decades. Because it is evident there exists high interest in the field, it is important to act on this interest and invest more resources in AI projects for drug discovery. Additionally, these new projects must avoid the challenges identified in current research and focus more on areas ready for innovation. Once these steps are taken, there will be more instances of successful drugs that can be pushed into clinical trials and eventually used to treat patients.

*Current Challenges.*

### 1.5.1. *Data Quality.*

One of the biggest challenges that faces the process of drug discovery that is AI-driven by poor data quality. Large and high-quality datasets are critical to achieving good results with AI models. However, data used in the process of drug discovery primarily come from different sources such as clinical trials, laboratory experiments, and public databases. These data are either noisy, incomplete, or biased; hence, AI models make inaccurate predictions based on them (Chan et al., 2019). Additionally, the proprietary nature of most pharmaceutical data means the amount of high-quality information available for training AI models is limited. This is further compounded by the fact that too many of the datasets that are available are biased toward drug candidates that have been successful, therefore skewing model predictions and reducing the ability of a model to generalize to novel compounds (Chan et al., 2019).

It will be necessary to standardize data collection, curation, and sharing. Efforts to establish large, open-access databases with uniform formats could significantly enhance the quality and diversity of data for AI training (Bender and Cortes-Ciriano, 2021). Besides, data augmentation, transfer learning, and synthetic data generation are other means that have been explored for enhancing the quality of data in an effort to reduce bias within AI-driven drug discovery (Chan et al., 2019).

1.6. *Model Generalization.*

One of the other critical challenges in AI-driven drug discovery is model generalization. Models created with AI are highly successful in making predictions of results within their learning scope but usually fail to generalize into new, unseen data. This may turn specifically problematic in the case of drug discovery, for its prime function is the prediction of behaviors of new compounds (Stokes et al., 2020). One common problem that really undermines the reliability of AI predictions is overfitting, where a model performs well on training data and then completely fails on new data (Chan et al., 2019). Regularization, cross-validation, and ensemble learning are some attempts at better generalization of models. All these methods can, to some extent, reduce overfitting of AI models and make them more robust. In addition, more diverse training datasets—that is, datasets with a wide variety of structures and biological targets—will further increase the generalizability of models at hand (Gorgulla et al., 2020). In AI-driven drug discovery, however, it remains a challenge to balance model complexity with generalizability (Stokes et al., 2020).

1.7. *Integration of ML in Existing Pipelines.*

Another challenge is the integration of ML into the existing drug discovery pipelines. The process of drug discovery is long and comprises many complicated steps: from target identification to lead discovery, preclinical testing, and clinical trials. All these steps require different kinds of data and areas of expertise, so it is hard to see how ML would seamlessly fit into the entire pipeline (Chan et al., 2019). Also, many traditional drug discovery processes are poorly suited for the iterative use of ML, in which models are being continuously updated by new data and feedback (Chan et al., 2019). It is thus incumbent to develop more flexible and adaptable drug discovery pipelines that can make use of ML at a variety of stages. This could, in certain instances, entail redesigning existing processes to be data-driven and iterative but also entails building strong collaboration between computational scientists and domain experts in biology and chemistry (Bender and Cortes-Ciriano, 2021). Moreover, the integrated development of platforms that bring together data management and ML with experimental validation into one environment could be another area of potential improvement in the integration of ML into drug discovery (Gorgulla et al., 2020).

1.2. *Lessons Learned from AI-Driven Drug Discovery.*

The success of AI-driven drug discovery projects has not come without challenges. One of the main messages that emerged from this was that data quality really matters, and AI models do have their ceiling of operations. Some of those AI models could just be successful only if there were very high quantities of quality data available to make predictions with accuracy. These models thus rely heavily on the training data used, making them prone to mistakes in the presence of biases or inaccuracies in that data. For instance, if the training set is biased for some specific type of molecule or target, it would make an AI model fail in cases of other types of molecules or targets. This would indicate that larger, more diverse datasets are necessary for such AI-driven drug discovery (Chan et al., 2019).

Another lesson is the necessity of balancing AI-driven approaches with traditional experimental methods. Even if AI could speed up the rate of the drug discovery process greatly, experimental validation eventually needs to take place. The experimentally obtained facts, in turn, must be validated by strong in vitro and in vivo assays to be assured of their accuracy and application in the real world. This hybrid approach of combining AI with the traditional methods has turned out to be the most successful way in drug discovery. For example, during the development of DDR1 kinase inhibitors, although the AI model quickly picked up candidates, these candidates still had to undergo thorough testing in the lab for their efficacy and safety (Zhavoronkov et al., 2019).

Moreover, the integration of AI into drug discovery has under- scored the importance of interdisciplinary collaboration. Successful AI-based projects, on the other hand, are often done in concert with computational scientists, biologists, and chemists working together. Such an approach is critical to making sure AI models interpret biological data correctly to provide meaningful predictions. For example, in the Schrodinger/Sanofi partnership, computational chemists worked in concert with biologists to ensure that the de- signed AI-generated compounds would display suitable biological activity. This collaboration was hence a kernel in the success of the project, underlined by a key concept in AI-driven drug discovery: bringing together the expertise from several disciplines. In addition, the use of AI in drug discovery has highlighted the importance of continuous learning and adaptation. The AI models learn and adapt more with the training on more datasets or more challenging tasks, and this goes hand in hand with constant updating and fine-tuning to achieve higher accuracy and utility out of them. That will demand continuous investment in Research and Development, an openness toward new technologies, and novel approaches. Companies that can do this will better succeed as AI-driven drug discovery will no doubt change quickly in the years to come (Chan et al., 2019).

Lastly, the scalability of AI approaches in drug discovery has proven both a strength and a challenge. On the one hand, A can handle huge data sets and scan vast chemical spaces; on the other, this comes at a considerable computational cost. Inevitably, this has created growing recognition of the need for better infrastructure of computing and the development of more efficient algorithms scaling effectively with the increased complexity of drug discovery projects (Gorgulla et al., 2020).

**Future Trends.**

1.1. *Quantum Computing.*

Quantum computing represents one of the bright areas of emerging trends in AI-driven drug discovery. Contrasted to classical computers representing information in the form of 0 or 1, quantum computers make use of qubits, which represent the information in both 0 and 1. This feature, called superposition, makes a quantum computer conduct many calculations all at one time. Another major feature of quantum computers is entanglement, allowing qubits to become correlated in such a way that one depends on the state of another, even when large distances exist between them. Due to these features, quantum computers are able to run such complex calculations unbelievably fast, thus taking a very significant advantage in molecular modeling and drug design (Gorgulla et al., 2020). This means simulating the behavior of molecules and how they act against biological targets has been computationally intensive traditionally in the process of drug discovery. These simulations normally pose problems in the complexity encountered by classical computers, in particular for large molecules or complex interactions. Quantum computing holds the potential to make a difference in this part of drug discovery by enabling simulations at a scale in terms of detail and speed unreachable for classical computers. This would hence drastically increase the rate of drug discovery by searching at great speed in the huge chemical space and simulating intricate biological systems more precisely and efficiently (Bender and Cortes-Ciriano, 2021).

Moreover, quantum computers can solve one of the most challenging tasks in drug discovery: the accurate prediction of drug-target interactions. Most classical computational methods are very often based on approximations that make them less accurate. Quantum computers can simulate quantum-mechanical systems immediately and provide much greater accuracy when it comes to understanding how drugs interact with their targets at an atomic level. It can result in the discovery of new drugs that were earlier ignored because classical computation is not capable enough to track their activities (Biamonte et al., 2017).

The integration of AI with quantum computing further amplifies these advantages, especially in drug discovery. AI algorithms can optimize quantum simulations by selecting the most relevant molecular structures to simulate, thereby reducing the computational load and improving the efficiency of quantum computing tasks. For instance, AI can be used to predict which molecular interactions are likely to be the most promising before they are subjected to quantum simulations, saving valuable computational resources. Additionally, AI-driven models can interpret the vast amounts of data generated by quantum simulations, identifying patterns and correlations that might not be immediately apparent through quantum computations alone (Biamonte et al., 2017). Quantum AI, the intersection of quantum computing and AI, represents a powerful tool for drug discovery. Quantum AI can be utilized to develop novel algorithms that are specifically designed for quantum hardware, leveraging the strengths of both quantum computing and AI. These algorithms can optimize the drug discovery process by rapidly screening vast libraries of compounds, predicting their properties, and identifying the most promising candidates for further testing. The synergy between AI and quantum computing could potentially overcome many of the limitations currently faced by classical computational methods, such as the inability to accurately model complex biological systems and the inefficiency of screening large chemical spaces (Perdomo-Ortiz et al., 2018).

Though quantum computing is at its emerging stage, over- whelming development in quantum algorithms for drug discovery is underway. For example, a group of scientists is currently working on utilizing quantum computers in simulating the quantum behavior of molecules that could have potential applications in more accurate predictions of drug-target interactions (Gorgulla et al., 2020). These quantum simulations encapsulate the full quantum mechanical nature of molecules, which is required for understanding complex phenomena like electron correlation and entanglement, very critical in chemical reactions and molecular stability (Bender and Cortes-Ciriano, 2021).

AI-driven approaches are also being developed to enhance the capabilities of quantum computing in drug discovery. For instance, ML algorithms are being trained to identify specific quantum states that are most relevant for drug discovery applications, which could significantly improve the accuracy of quantum simulations. Furthermore, AI can assist in the design of quantum algorithms by identifying the most efficient computational pathways for simulating drug interactions, thereby reducing the time and resources required to achieve accurate results (Cova et al., 2022).

Quantum computing may extremely raise the accuracy of simulations and drastically reduce times for screening potential drug candidates. Where years are taken to screen large chemical libraries with classical computers, a quantum computer would do it within a few days or even hours. Such acceleration in computational efforts can really move the drug development timelines up to make new therapies available to patients faster (Perdomo-Ortiz et al., 2018).

It is envisioned that by the near future, improving quantum hardware will further place quantum computing at a more prime position in AI-bolstered drug discovery. In the case of advanced hardware, if the number of qubits is large and error correction is much better, then practical applications of quantum computing in drug discovery become increasingly viable. This will most likely result in breakthroughs for new therapies, especially in such com- plex diseases where old conventional methods have not been able to find adequate treatments. It can open up whole new avenues for quantum computing and AI ntegrated with ML to achieve so much more personalized medicine where therapies have a basis upon the individual genetic makeup of a patient for enhanced efficacy with reduced side effects (Biamonte et al., 2017).

With the maturing of quantum computing technology, it is likely to play an increasing role in drug discovery, from drug candidate design and screening to drug manufacturing process optimization and the development of new materials for drug delivery. Quantum computing applied to drug discovery has a shining future ahead. It possesses a huge potential in the pharmaceutical sector, leading to the revolution of quick and effective medicine development (Cova et al., 2022).

1.2. *Federated Learning.*

Federated learning is one such emerging trend that would eventually revolutionize AI-driven drug discovery. In traditional ML, data from multiple sources get centralized in one location for training models.

The above approach causes a few concerns regarding data privacy and security, particularly within the pharmaceutical industry, where any sort of proprietary data are highly sensitive (Stokes et al., 2020). Federated learning resolves these concerns by training AI models across multiple decentralized datasets without sharing the raw data (Gu et al., 2023).

In such a federated learning framework, each organization would be able to train AI models on its local data and only share model updates—e.g., gradients—with a central server. These updates can then be aggregated on the server without touching the raw data to improve the global model (Gorgulla et al., 2020). This approach will retain the privacy and security of an organization's data while providing collaboration toward AI-driven drug discovery (Bender and Cortes-Ciriano, 2021). It may also entail that due to the very diverse sources—like many pharmaceutical companies, research institutions, and healthcare providers—federated learning would mean more robust and generalizable models (Stokes et al., 2020).

1.3. *AI for Personalized Medicine.*

Another area in which AI is very likely to have enormous impacts is personalized medicine, tailored to each patient as a function of genetic background, environmental factors, and lifestyle. Other than the many promises of AI-driven drug discovery, it can basically speed up the progress of personalized therapies by large-scale genomic and clinical data analyses for biomarker identification, prediction of patient responses against drugs, and designing a treatment plan accordingly (Chan et al., 2019).

One of the major challenges to be faced in personalized medicine is complexity and heterogeneity in biological data. AI models that have the potential to merge data from, say, genomics, proteomics, and electronic health records, and analyze it, would be critical in the advancement of personalized medicine (Bender and Cortes-Ciriano, 2021). More specifically, AI-driven in silico simulations of drug effects in individual patients could provide the means to have more precise and effective treatments and reduce trial-and-error approaches so common in clinical practice today (Gorgulla et al., 2020). In development, AI will probably play a central role in the next generation of therapies that are truly personalized.

**Opportunities for Innovation.**

1.3.1. *Gaps in Current Research.*

Despite the progress made in AI- driven drug discovery, there are several gaps in current research that present opportunities for innovation. The most prominent gap is the lack of standardized high-quality datasets to train AI models. While some efforts have been initiated toward the creation of open- access databases, further work must be directed at ensuring such datasets are representative and unbiased, with no lacunae (Chan et al., 2019). It's not only about the volume; it's also related to the quality improvement of data through better curation, annotation, and validation (Chan et al., 2019).

Another gap is in the understanding of the intrinsic biological mechanisms that drive drug-target interactions. While AI models can make accurate predictions based on patterns in data, often lacking interpretability makes it hard to understand what the biological basis for these predictions is (Bender and Cortes-Ciriano, 2021). More interpretable AI models that aid in understanding the mechanism of action would increase the chances of effective therapy for any drug and significantly lower the risk of unexpected side effects (Bender and Cortes-Ciriano, 2021). It will boil down to closer collaboration between the AI researcher and the biologist in the incorporation of domain knowledge into AI models.

1.3.2. *Areas Ripe for Innovation.*

There are many areas within the AI-driven drug discovery process that are primed for innovation. Among them is the development of hybrid models combining AI techniques with other, more traditional computational approaches, like molecular dynamics simulations and quantum chemistry. Hybrid models may be developed that bring out the strengths of both approaches in making improved and more reliable predictions about drug behavior (Gorgulla et al., 2020). For instance, AI can be used in the high-throughput screening of large compound libraries while molecular dynamics simulations inform about detailed drug-target interactions (Bender and Cortes-Ciriano, 2021).

One such other area of innovation would be the employment of AI in designing new drug delivery systems. Traditional drug delivery is often riddled with issues related to low or no bioavailability, off-target impacts, or even eliciting drug resistance. AI can design more effective delivery systems by optimizing these factors: drug formulation, release kinetics, and targeting strategies (Vora et al., 2023). AI could be applied in designing nanoparticles, for instance, so that specific tissues or cells are targeted for drug delivery with a much greater degree of accuracy, therefore reducing side effects and increasing therapeutic outcomes (Chan et al., 2019).

The other highly promising area is AI-driven drug repurposing. Since a large part of drug development time and cost gets reduced through the process of drug repurposing, which involves finding new therapeutic uses for existing drugs, it become attractive. AI models that can compare molecular structure, mode of action, or clinical outcome would identify potential new indications for the approved drugs (Stokes et al., 2020). Thus, this might lead to rapid discovery for a myriad of disease treatments (Bender and Cortes-Ciriano, 2021).

1.3.3. *Collaborative Innovation.*

Finally, in the future, collaborative innovation among academia, industry, and regulatory bodies will be necessary to advance drug discovery driven by AI. The contribution from academia is necessary for leading-edge research and new methodologies; industries can contribute with the resources and infrastructures that allow the scaling of innovations (Chan et al., 2019). The regulatory bodies will give the guidelines and standards that will govern the application of AI in drug discovery, such that these technologies are safe, efficient, and ethical (Chan et al., 2019). On the other hand, as much as AI-driven drug discovery comes with a host of formidable challenges, it also opens up many opportunities for innovation. In this line, solving challenges in data quality, model generalization, and integration into existing pipelines, and embracing quantum computing, federated learning, and personalized medicine, among many other emerging trends, this field can keep evolving and create new possibilities in drug discovery. AI can reach its full potential to fundamentally change this area only through collaborative efforts across disciplinary and sectoral lines (Chan et al., 2019).

**Conclusion**

AI-driven drug discovery has transformed the pharmaceutical industry, offering new possibilities for the rapid and cost-effective development of novel therapeutics. Therefore, successful AI- driven projects represent collaborations between pharmaceutical companies and AI tech firms in proving that this is the potential these technologies have for drug discovery. These successes, however, have pointed to the requirement of high-quality data, interdisciplinary collaboration, and a balanced approach that puts a premium on both AI and traditional experimental methods. As AI continues into the future, its development will be critically key to the future of drug discovery, opening up many more treatments and therapies for a wide range of diseases. While ML is promising, it's not perfect. One major issue is the availability of large, high-quality datasets. Available data are often incomplete, noisy or biased, which can lead to unreliable predictions. Additionally, many ML models, particularly those based on DL, are often described as "black boxes." These ML models can make high quality predictions, it's extremely difficult to figure out exactly how they come to those results. This lack of transparency can be a significant barrier from ML being utilized more in the drug discovery field (Fleming, 2018).

Overfitting is another significant issue in current ML models which occurs when a model is too closely fitted to its training data and doesn't respond properly to new, unseen data. In the context of drug discovery, overfitting could be a significant issue as a drug can show promise in initial tests, but fail under real world scenarios in which there are more unique variables to factor in. Addressing these challenges is crucial in order to make ML more widespread and useful in the drug discovery field (Cohen et al., 2021; Stokes et al., 2020).

While ML models in drug discovery still pose many issues, these issues can bring forth new opportunities. For example, the integration of multi-omics data, which is information from genomics, proteomics, and metabolomics, could provide a more detailed understanding of disease mechanisms which in turn can lead to the identification of new drug targets (Du et al., 2024). Additionally, advances in natural language processing (NLP) as well as reinforcement learning are allowing for the automation of more aspects of the drug discovery process, ranging from mining scientific literature to optimizing chemical synthesis pathways (Schneider et al., 2020).

**Author Contributions**

**Khartik Uppalapati**: Writing—original draft, writing—review & editing, conceptualization, methodology, formal analysis, data curation. **Eeshan Dandamundi:** Writing—review & editing, conceptualization, methodology, formal analysis, data curation. **S. Nick Ice**: Writing—review & editing, conceptualization, methodology, formal analysis, data curation. **Gaurav Chandra -** Writing—review & editing, conceptualization, methodology, formal analysis, data curation. **Kristen Bischof -** Writing—review & editing, conceptualization, methodology, formal analysis, data curation. **Christian L. Lorson**: Writing—review & editing. **Kamal Singh**: Writing—original draft, writing—review & editing, conceptualization, methodology, formal analysis, data curation, supervision, funding acquisition.

**Acknowledgments**

K. Singh acknowledges the computation facilities of the Molecular Interactions Core at the University of Missouri, Columbia, MO 65212, and funding support from the University of Missouri.

**Ethics declarations**

**Legend**

Figure 1. Plot highlighting the popularity of AIe in the Antibiotic Discovery Process since 1975. A net increase in the popularity of 10,000 Peer-Articles from the PubMed database containing any given combination of two or more search terms from each of two keyword sets containing the terms, "Artificial Intelligence, Deep Learning, and Machine Learning" as well as "Antibiotic Discovery, Antibiotics, and Drug Discovery" respectively can be seen be seen over the years.

**Abbreviations**

Machine Learning – ML

Artificial Intelligence – AI

Computer-Aided Drug Discovery – CADD

Quantitative Structure-Activity Relationship – QSAR

Deep Learning – DL

Convolutional Neural Network – CNN

Graph Neural Network – GNN

Antimicrobial resistance – AMR

Support Vector Machine – SVM

Random Forest – RF

Recurrent Neural Networks (RNNs)

Long Short-Term Memory – LSTM

Generative Adversarial Network – GAN

Variational Autoencoders – VAE

Reinforcement Learning – RL

Simplified Molecular Input Line Entry System – SMILES

Cost-Sensitive Learning – CSL

Genome Wide Association Studies – GWAS

genome-scale models – GEM

**Table 1.** Compiled list of the major Machine Learning datasets for drug discovery. Each of the 18 represented datasets is shown with an access link, description, and brief explanation of its applications. The selected datasets are diverse in their applications at each stage of the drug discovery process.

| Dataset | Link | Definition | Application |
| --- | --- | --- | --- |
| QM9 | http://quantummachine.org/datasets | Provides quantum chemical properties for small organic molecules | Machine learning models learn the patterns in QM9 and generate small drug-like molecules |
| Tox21 | http://bioinf.jku.at/research/DeepTox/tox21.html | Training and test samples for chemical compounds | Helps ML predict drug toxicity levels |
| BACE | https://drugdesigndata.org/about/grand-challenge-4/bace | Quantitative and qualitative binding results | Helps ML predict binding affinity of drugs |
| BBBP | https://www.kaggle.com/datasets/priyanagda/bbbp-smiles | Binary labels for blood-brain-barrier penetration (BBBP) | Drugs ML identifies for neurological conditions must have BBBP capabilities |
| PDD | https://www.ncbi.nlm.nih.gov/pathogens | Previously sequenced disease isolates | Allows ML to identify significant genes associated with disease stress response |

| HIV | https://wiki.nci.nih.gov/display/NCIDTPdata/AIDS+Antiviral+Screen+Data | Screening results for inhibiting HIV replication | ML can predict anti-HIV activity |
|---|---|---|---|
| SIDER | http://sideeffects.embl.de/ | Side effects of marketed medicines | ML can predict the side effects of drugs |
| ClinTox | https://huggingface.co/datasets/zpn/clintox | Comparison of approved and non-approved FDA drugs | ML can predict whether drugs will be FDA approved |
| ToxCast | https://www.epa.gov/comptox-tools/exploring-toxcast-data | Predicts potential toxicity of chemical compounds | Used to train ML models to predict drug toxicity |
| BindingDB | https://registry.opendata.aws/binding-db | Binding affinity measurements | Helps ML predict drug-target interactions |
| ESOL | https://togodb.biosciencedbc.jp/togodb/view/esol#en | Water solubility of samples | Helps ML predict aqueous solubility of drugs |
| FreeSolv | https://pubmed.ncbi.nlm.nih.gov/24928188/ | Calculated hydration-free energies of small molecules | Helps ML predict drug solvation properties |
| MUV | https://pubs.acs.org/doi/10.1021/ci8002649 | Validates virtual screening methods | Benchmarks ML scoring functions |
| DAVIS-DTA | https://www.nature.com/articles/nbt.1990 | Interactions between kinases and kinase inhibitors | Helps ML identify novel inhibitors |
| KIBA | https://www2.helsinki.fi/sites/default/files/atoms/files/kiba.zip | Integrated drug-target bioactivity matrix | Helps ML predict drug-target interactions |
| Guacamol | https://benevolent.ai/guacamol | Assesses deep learning and neural generative models for drug discovery | Benchmarks models for de novo drug design |
| ZINC | https://zinc.docking.org/ | 3D molecules of chemical compounds | Used for docking and ML virtual screening to identify inhibitors |
| DrugBank | https://go.drugbank.com/ | Information on drugs and drug targets | ML can rapidly screen compounds for potential |

Table 2. Equation Symbol Notations for SPARROW. List of the notations used to develop the SPARROW algorithm. Documentation of the purpose of various components comprising SPARROW e.g. nodes, cycles, indices, compound nodes, and decision variables.

| Symbol | Definition |
|---|---|
| $j$ | An index referring to a reaction node |
| $i$ | An index referring to a compound node |
| $R$ | Set of reaction node indices |
| $C$ | Set of compound node indices |
| $c_j$ | Decision variable defining whether compound node $j$ is selected |
| $r_i$ | Decision variable defining whether reaction node $i$ is selected |
| $S$ | Set of dummy reaction node indices |
| $P_{i\ or\ j}$ | Set of parent nodes for the node corresponding to index $i\ or\ j$ |
| $y$ | A cycle in a retrosynthetic graph |

Figure 1. Plot highlighting the popularity of Artificial Intelligence in the Antibiotic Discovery Process since 1975. A net increase in the popularity of 10,000 Peer-Articles from the PubMed database containing any given combination of two or more search terms from each of two keyword sets containing the terms, "Artificial Intelligence, Deep Learning, and Machine Learning" as well as "Antibiotic Discovery, Antibiotics, and Drug Discovery" respectively can be seen be seen over the years.

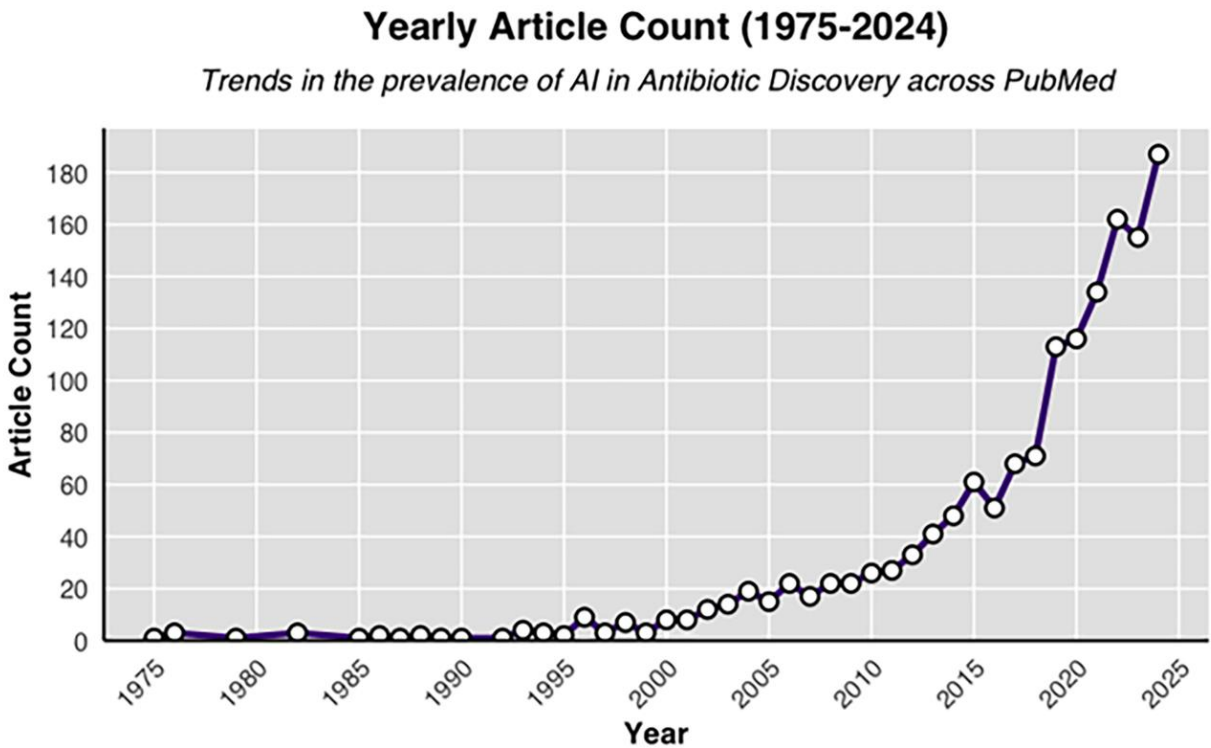